# Learning by Transduction

A. Gammerman, V. Vovk, V. Vapnik
Department of Computer Science
Royal Holloway, University of London
Egham, Surrey TW20 0EX, UK
{alex,vovk,vladimir}@dcs.rhbnc.ac.uk

## Abstract

We describe a method for predicting a classification of an object given classifications of the objects in the training set, assuming that the pairs object/classification are generated by an i.i.d. process from a continuous probability distribution. Our method is a modification of Vapnik's support-vector machine; its main novelty is that it gives not only the prediction itself but also a practicable measure of the evidence found in support of that prediction. We also describe a procedure for assigning degrees of confidence to predictions made by the support vector machine. Some experimental results are presented, and possible extensions of the algorithms are discussed.

## 1 THE PROBLEM

Suppose labeled points $(x_i, y_i)$ $(i = 1, 2, \ldots)$, where $x_i \in \mathbb{R}^n$ (our objects are specified by $n$ real-valued attributes) and $y_i \in \{-1, 1\}$, are generated independently from an unknown (but the same for all points) probability distribution. We are given $l$ points $x_i$, $i = 1, \ldots, l$, together with their classifications $y_i \in \{-1, 1\}$, and an $(l+1)$th unclassified point $x_{l+1}$. How should it be classified? (This is a problem of *transduction*, in the sense that we are interested in the classification of a particular example rather than in a general rule for classifying future examples; for further discussion of transduction, see Section 6.)

A natural and well-known approach is Vapnik's [7] method of support vector (SV) machines. The SV method works very well in practice, but unfortunately no practicable estimates of the accuracy of its predictions are known if our only information is $l$ classified points and one unclassified point. The most relevant, in this context, theorem from [7] (Theorem 5.2) says that the probability of misclassifying the $(l+1)$th point is at most

$$\frac{\mathbf{E}(\text{number of support vectors among } x_1,\ldots,x_{l+1})}{l+1}, \quad (1)$$

where the points $x_1, \ldots, x_{l+1}$ are generated independently from the underlying distribution $P$; support vectors are defined in Section 5 below. To apply this theorem we need to know the probability distribution $P$, while the only information we do know is

$$(x_1, y_1), \ldots, (x_l, y_l), x_{l+1}.$$

Clearly this is not sufficient to estimate the expectation in (1).

**Remark 1** Dawid [2] distinguishes between nominal and stochastic inference; in our present context nominal inference is the prediction itself and stochastic inference is some assertion about the accuracy of this prediction. To use this terminology, the SV method provides only nominal but no stochastic inference. (Of course, since the SV method is being actively developed, the situation is likely to change in the future.)

## 2 PREDICTING WITH CONFIDENCE

Now we briefly describe, following [4], our transductive algorithm, putting off its substantiation until Section 5. We consider two pictures in the space $\mathbb{R}^n$: both pictures contain $(l+1)$ points (the $l$ points in the training set and one point to be classified), the points in the training set are classified as before, and the only difference between the pictures is the classification of the $(l+1)$th point; in the $-1$-*picture* that point is classified as $-1$ and in the $1$-*picture* it is classified as $1$. It can be proven that the $(l+1)$th point will be a support vector in at least one of the pictures. Let $SV(1)$ (resp. $SV(-1)$) be the set of indices of support vectors in the $1$-picture (resp. $-1$-picture); we let $\#A$ stand for the cardinality of set $A$. Our algorithm gives the following predictions and "incertitudes":

- 1 if

$$((l+1) \in \text{SV}(-1)) \ \& \ ((l+1) \notin \text{SV}(1))$$

or

$$((l+1) \in \text{SV}(-1) \cap \text{SV}(1))$$
$$\& \ (\#\text{SV}(-1) < \#\text{SV}(1));$$

the incertitude of this prediction is

$$\frac{\#\text{SV}(-1)}{l+1};$$

- $-1$ if

$$((l+1) \in \text{SV}(1)) \ \& \ ((l+1) \notin \text{SV}(-1))$$

or

$$((l+1) \in \text{SV}(-1) \cap \text{SV}(1))$$
$$\& \ (\#\text{SV}(1) < \#\text{SV}(-1)),$$

with incertitude

$$\frac{\#\text{SV}(1)}{l+1};$$

- any prediction if

$$((l+1) \in \text{SV}(-1) \cap \text{SV}(1))$$
$$\& \ (\#\text{SV}(-1) = \#\text{SV}(1))$$

with incertitude

$$\frac{\#\text{SV}(-1)}{l+1} = \frac{\#\text{SV}(1)}{l+1}.$$

The interpretation of incertitude is as follows: failure of a prediction of incertitude $\mu$ is as likely as a win of $\pounds\frac{1}{\mu}$ on a £1 ticket in a fair lottery. Exact definitions will be given below.

Our method works well (gives predictions of small incertitude or, in other words, confident predictions) when the number of support vectors is small in both pictures. When applying the SV method, one hopes that this assumption will be satisfied; in the experiments presented in Vapnik [7] (footnote 4 on p. 131) the support vectors constituted 3% to 5% of the data set. Our experiments (see Section 7) have typically given incertitudes 0.05–0.10. It is often easier to think in terms of confidences rather than incertitudes; we define *confidence* to be $1-I$, $I$ being incertitude. (So incertitudes 5–10% correspond to confidences 90–95%.)

The transductive algorithm described above was designed to optimize the confidence. In our computer experiments, however, we found that its performance (measured by the number of mistakes made) is not as good as the performance of the standard SV algorithm (see Section 7 below). Of course, this finding does not mean that the standard algorithm is definitely better, because in some applications confidence might be even more important than performance; but it does make it desirable to introduce some measure of confidence for the standard SV algorithm. The following procedure provides a measure of confidence for *any* prediction algorithm (and in the case of our transductive algorithm it gives the same confidences).

As before, we have two pictures, the $-1$-*picture* and the $1$-*picture*. Let $\hat{y}$ be the prediction for $y_{l+1}$ made by the given prediction algorithm. The incertitude of this prediction is defined to be

1. $\dfrac{\#\text{SV}(-\hat{y})}{l+1}$, if $(l+1) \in \text{SV}(-\hat{y})$;

2. $\infty$, otherwise.

(The interpretation of incertitude is the same as before: failure of a prediction of incertitude $\mu$ is as likely as a win of $\pounds\frac{1}{\mu}$ on a £1 ticket in a fair lottery.) For the SV algorithm, only possibility 1 can realize: it is always true that $(l+1) \in \text{SV}(-\hat{y})$. So in this most interesting for us case our procedure of assigning confidence is extremely simple: the confidence is just $1 - \frac{\#\text{SV}(-\hat{y})}{l+1}$.

In our experiments we found that a very important role is played by what we call the "possibility" of the data set; this is discussed in Sections 4 and 7 below.

## 3   MEASURES OF IMPOSSIBILITY

In this section we introduce notions which will enable us to define the notion of incertitude; our exposition will partly follow [9] and [10].

Let $\Omega$ be some sample space (a typical sample space is the set of all sequences $(x_1, \ldots, x_{l+1})$ of $l+1$ points in the Euclidean space $x_i \in \mathbb{R}^n$ with their classifications $y_i \in \{-1, 1\}$, $i = 1, \ldots, l+1$, equipped with the usual $\sigma$-algebra). If $P$ is a probability distribution in $\Omega$, a $P$-*measure of impossibility* is defined to be a non-negative measurable function $p: \Omega \to \mathbb{R}$ such that

$$\int_\Omega p(\omega) P(d\omega) \leq 1. \qquad (2)$$

This is our explication of the notion of lottery; we visualize $P$ as the randomizing device used for drawing lots and $p(\omega)$ as the value of the prize won by a particular ticket when $P$ produces $\omega$. Notice that we do not exclude "fair" lotteries which satisfy (2) with an equality sign (which means that all proceeds from selling the tickets are redistributed in the form of prizes), though in real lotteries the left-hand side of (2) is usually much less than 1.

By Chebyshev's inequality, $p$ is large with small probability: for any constant $C > 0$,

$$P\{\omega \in \Omega : p(\omega) \geq C\} \leq \frac{1}{C}.$$

This confirms our intuition that if $p$ is chosen in advance and we believe that $P$ is the true probability distribution generating the data $\omega \in \Omega$, then it is hardly possible that $p(\omega)$ will turn up large.

**Remark 2** The notion of a "critical region" used in the theory of testing statistical hypotheses is essentially a special case of our notion of a measure of impossibility: a subset $A \subseteq \Omega$ of the sample space of a small probability $\delta = P(A)$ is identified with the $P$-measure of impossibility

$$p(\omega) = \begin{cases} 1/\delta, & \text{if } \omega \in A, \\ 0, & \text{otherwise.} \end{cases}$$

If $\mathcal{P}$ is a family of probability distributions, we define a $\mathcal{P}$-measure of impossibility to be a function which is a $P$-measure of impossibility for all $P \in \mathcal{P}$. Most of all we are interested in the $\mathcal{C}^m(Z)$-measures of impossibility, where $Z$ is a measurable space, $m$ is a positive integer (the sample size), and $\mathcal{C}^m(Z)$ stands for the set of all product distributions $P^m$ in $Z^m$, $P$ running over the continuous distributions in $Z$. Our interpretation of this definition is as follows: if $p$ is a $\mathcal{C}^m(Z)$-measure of impossibility and $z_1, \ldots, z_m$ are generated independently from a continuous distribution, it is hardly possible that $p(z_1, \ldots, z_m)$ is large (provided $p$ is chosen before the data $z_1, \ldots, z_m$ are generated).

Now we shall introduce an important subclass of the $\mathcal{C}^m(Z)$-measures of impossibility. A non-negative measurable function $p : Z^m \to \mathbb{R}$ is a *permutation measure of impossibility* if, for any sequence $z_1, \ldots, z_m$ in $Z^m$,

- $p(z_1, \ldots, z_m) = \infty$ if $z_i = z_j$ for some $i \neq j$;
- $\frac{1}{m!} \sum_\pi p(z_{\pi(1)}, \ldots, z_{\pi(m)}) = 1$ (the sum is over all permutations $\pi$ of the set $\{1, \ldots, m\}$), if all elements of the set $\{z_1, \ldots, z_m\}$ are different.

It is obvious that every such $p$ is indeed a $\mathcal{C}^m(Z)$-measure of impossibility.

## 4  GENERAL SCHEME

First we describe our task. We fix a training set size $l$ (a positive integer) and an attribute space $X$ (an arbitrary measurable space). Put $Z = X \times \{-1, 1\}$ ($Y = \{-1, 1\}$ is our *label space*). We are given a sample $z_1, \ldots, z_l$ of classified examples, $z_i = (x_i, y_i) \in Z$, $i = 1, \ldots, l$, and one unclassified example $x_{l+1} \in X$; $(x_i, y_i)$ are assumed to be generated independently from some unknown probability distribution $P$ in $Z$. Our goal is to predict the classification $y_{l+1} \in \{-1, 1\}$ of $x_{l+1}$.

Our algorithm for doing so is as follows. First we choose a permutation measure of impossibility $p : Z^{l+1} \to \mathbb{R}$. After observing $z_1, \ldots, z_l, x_{l+1}$ we calculate two values:

$$\mu_{-1} = 1/p(z_1, \ldots, z_l, (x_{l+1}, -1))$$

and

$$\mu_1 = 1/p(z_1, \ldots, z_l, (x_{l+1}, 1)).$$

Then we predict with $\arg\max \mu$ (i.e., predict with 1 if $\mu_{-1} < \mu_1$, with $-1$ if $\mu_{-1} > \mu_1$, and predict arbitrarily if $\mu_{-1} = \mu_1$); the *incertitude* of our prediction is

$$\mu = \min(\mu_{-1}, \mu_1)$$

(and our *confidence* in our prediction is $1 - \mu$). The interpretation of this measure of our incertitude is that our prediction is right unless a £1 ticket wins £$\frac{1}{\mu}$ in a lottery; if $\mu$ is small, we can be pretty sure that our prediction is correct.

Notice that Chebyshev's inequality implies

$$P\{\mu \leq \epsilon \ \& \ \text{prediction is wrong}\} \leq \epsilon,$$

for any constant $\epsilon > 0$ and any distribution $P$.

The quality of data is given by the *possibility*

$$\max(\mu_{-1}, \mu_1).$$

If this value is small, $p(z_1, \ldots, z_{l+1})$ is guaranteed to be big no matter which $y_{l+1}$ will turn up; therefore, such data are hardly possible, and our experiments have shown that the quality of prediction for such data is typically very poor. Notice that the notion of possibility does not depend (unlike confidence) on the prediction algorithm used and is a property of the data. We will usually truncate the value of possibility reporting 1 in the case where it exceeds 1.

The prediction algorithm described above optimizes the confidence of the predictions made. If, however, we have already decided on the algorithm to be used, we can associate a measure of incertitude with the algorithm's predictions as follows: the incertitude of a prediction $\hat{y}$ for $y_{l+1}$ is

$$\mu_{-\hat{y}}.$$

The interpretation of this measure of incertitude is analogous to what we had before: the prediction $\hat{y}$ is correct unless a £1 ticket wins £$\frac{1}{\mu}$ in a lottery.

## 5  SV IMPLEMENTATION

In the previous section we described a general prediction scheme (in particular, this scheme covers Fraser's [3] procedure of nonparametric prediction); in this section we shall consider a powerful implementation of this general scheme.

To begin with, we briefly describe one of the possible definitions of support vectors (see Cortes and Vapnik [1], Sections 3 and A.2, or Vapnik [7]). This definition is usually applied not to the original data but to their images under some, often non-linear, transformation. In this paper, we shall always assume that this transformation is identical; extension of our results to the general case is trivial.

Let our data be $((x_1, y_1), \ldots, (x_{l+1}, y_{l+1}))$, where $x_i \in \mathbb{R}^n$ and $y_i \in \{-1, 1\}$, $i \in \{1, \ldots, l+1\}$ (our notation $l+1$ for the sample size is chosen for agreement with the rest of the paper). Examples with $y_i = 1$ (resp. $y_i = -1$) will be called *positive* (resp. *negative*). Consider the quadratic optimization problem

$$\Phi(w, \xi) = \frac{1}{2}(w \cdot w) + C\left(\sum_{i=1}^{l+1} \xi_i^2\right) \to \min \quad (3)$$

$$\left(w \in \mathbb{R}^n,\ \xi = (\xi_1, \ldots, \xi_{l+1}) \in \mathbb{R}^{l+1}\right),$$

where $C$ is an *a priori* fixed positive constant, subject to the constraints

$$y_i((w \cdot x_i) + b) \geq 1 - \xi_i, \quad i = 1, \ldots, l+1, \quad (4)$$

$$\xi_i \geq 0, \quad i = 1, \ldots, l+1. \quad (5)$$

**Lemma 1** *Quadratic optimization problem (3) with constraints (4) and (5) has a unique solution provided the sample $(x_1, y_1), \ldots, (x_{l+1}, y_{l+1})$ contains both positive and negative examples.*

**Proof** First, it is clear that a solution exists. Let

$$\left(w^{(1)}, b^{(1)}, \xi^{(1)}\right),\ \left(w^{(2)}, b^{(2)}, \xi^{(2)}\right)$$

(where $\xi^{(j)} = \left(\xi_1^{(j)}, \ldots, \xi_{l+1}^{(j)}\right)$, $j = 1, 2$) be any two solutions. Their mixture

$$\left(\frac{w^{(1)} + w^{(2)}}{2}, \frac{b^{(1)} + b^{(2)}}{2}, \frac{\xi^{(1)} + \xi^{(2)}}{2}\right)$$

will satisfy constraints (4) and (5) and, because of the strict convexity of the functional $\Phi(w, \xi)$, will provide a smaller value for this functional unless $w^{(1)} = w^{(2)}$ and $\xi^{(1)} = \xi^{(2)}$. Therefore, $w$ and $\xi$ are determined uniquely. If the minimum is attained at both $b = b_1$ and $b = b_2$ and $b_1 \neq b_2$, then all $\xi_i$ are zero (otherwise, at least one of the inequalities (4) with $\xi_i > 0$ and $b \in \{b_1, b_2\}$ would be strict and so the minimum in (3) would not be attained); however, in the separable case $b_1 \neq b_2$ is clearly impossible. □

We define a *support vector* to be any $(x_i, y_i)$ for which the corresponding inequality in (4) holds as equality. Consider a sample $(x_1, y_1), \ldots, (x_l, y_l)$ and one unclassified example $x_l$. As before, we consider the 1-picture, where $y_{l+1} = 1$, and the $-1$-picture, where $y_{l+1} = -1$. The most important, for our purposes, property of support vectors is the following.

**Lemma 2** *If the sample $(x_1, y_1), \ldots, (x_l, y_l)$ contains positive and negative examples, $z_{l+1}$ is a support vector in at least one of the two pictures.*

**Proof** This immediately follows from Lemma 1 and the simple observation that an inequality of type (4) cannot be strict (with $\xi_i = 0$) for both $y_i = 1$ and $y_i = -1$. □

We define a permutation measure of impossibility $p$ by

$$p(z_1, \ldots, z_{l+1}) = \begin{cases} \frac{l+1}{\#SV(z_1, \ldots, z_{l+1})}, & \text{if } z_{l+1} \in \text{SV}, \\ 0, & \text{otherwise,} \end{cases} \quad (6)$$

where $SV(z_1, \ldots, z_{l+1})$ are the support vectors in the set $\{z_1, \ldots, z_{l+1}\}$ and $z_i = (x_i, y_i)$, $i = 1, \ldots, l+1$. (We were assuming that $z_i$ are all different; if not, $p(z_1, \ldots, z_{l+1}) = \infty$ by definition.)

Now we can apply the general scheme of Section 4. By Lemma 2, we have three possibilities:

1. $z_{l+1}$ is a support vector only in $-1$-picture;
2. $z_{l+1}$ is a support vector only in 1-picture;
3. $z_{l+1}$ is a support vector in both pictures.

Let $\delta_{-1}$ be the fraction of the support vectors in the $-1$-picture and $\delta_1$ in the 1-picture; we will assume that $\delta_{-1}$ and $\delta_1$ are small (as already mentioned, this is typically the case). In cases 1 and 2, the incertitude of our prediction is

$$\min(\delta_{-1}, \delta_1); \quad (7)$$

in other words, we make a confident prediction. In case 3, whatever $y_{l+1}$ turns up, our permutation measure of impossibility $p$ will take a large value (at least $\frac{1}{\max(\delta_{-1}, \delta_1)}$), and so this case is hardly possible.

**Remark 3** Even in case 3 our algorithm still gives a confident prediction (assuming that $\delta_{-1}$ and $\delta_1$ are small), which looks counterintuitive. A. P. Dawid suggested quoting both $\min(\delta_{-1}, \delta_1)$ and $\max(\delta_{-1}, \delta_1)$ as the stochastic inference.

Notice that (7) is the incertitude of the prediction $\arg\max_y \delta_y$ in case 3 as well. This justifies the algorithm described in the Introduction.

We can see that Vapnik's SV method provides measures of impossibility that are especially well-suited to the scheme of Section 4. The reason why this is so is that:

- there are usually few support vectors;
- $z_{l+1}$ is a support vector in at least one of the pictures.

**Remark 4** It is clear that the above argument will hold if we replace "support vectors" by "essential support vectors", the latter notion being defined as follows. A vector $(x_j, y_j)$, $j \in \{1, \ldots, l+1\}$, is an *essential support vector* if the value of the optimization problem (3)–(5) does not change after deleting the term $\xi_j^2$ from the sum in (3) and deleting the constraints in (4) and (5) corresponding to $i = j$. The following example shows that these two notions (support vectors and essential support vectors) are indeed different. Consider the set $((x_1, y_1), \ldots, (x_{100}, y_{100}))$ of 100 classified examples in the plane defined as

$$x_i = (i, -1),\ y_i = -1,\quad i = 1, \ldots, 50,$$

$$x_i = (i - 50, 1),\ y_i = 1,\quad i = 51, \ldots, 100.$$

Here we have 100 support vectors and no essential support vectors.

We omit the derivation of the procedure of assigning confidences to the predictions made by the SV machine (see the end of Section 2) from the general procedure described at the end of Section 4.

## 6   TRANSDUCTION AND INDUCTION

"Transduction" is inference from particular to particular; for the problem of pattern recognition, it means that, given the classifications $y_i$, $i = 1, \ldots, l$, of the $l$ points $x_1, \ldots, x_l$ in the training set, we are only trying to guess the classifications of the $k$ points $x_{l+1}, \ldots, x_{l+k}$ in the test set. In the main part of this paper we only consider the case $k = 1$, though our methods can be easily extended to the case $k > 1$ (see Subsection 8.3 below).

In this section we are interested in the inductive implications of our procedure of transductive inference (with $k = 1$); recall that inductive inference, for our problem, requires that, given the training set $z_1, \ldots, z_l$, we should work out a general rule for classifying a future object $x$ as $-1$ or $1$. It is clear that our procedure describes such a general rule implicitly, and what we are interested in are the explicit aspects of this rule.

Let us solve the quadratic optimization problem

$$\frac{1}{2}(w \cdot w) + C \left( \sum_{i=1}^{l} \xi_i^2 \right) \to \min \quad \left( w \in \mathbb{R}^n,\ \xi \in \mathbb{R}^l \right),$$

$$y_i((w \cdot x_i) + b) \geq 1 - \xi_i,\quad \xi_i \geq 0,\quad i = 1, \ldots, l,$$

which is an analogue of (3)–(5) for the training set.

Let the unique (see Lemma 1) solution to this problem be $(w^*, b^*, \xi^*)$, and let the number of support vectors be $N$. We shall say that $x$ is a $y$-point, $y \in \{-1, 1\}$, if

$$y((w^* \cdot x) + b^*) > 1.$$

It is easy to see that our method will always predict $y$ for a $y$-point with incertitude $\frac{\#\mathrm{SV}(-y)}{l+1}$ (recall that $\#\mathrm{SV}(-y)$ is the number of support vectors in the $-y$-picture); therefore, if the fraction $\frac{\#\mathrm{SV}(-y)}{l+1}$ of support vectors in the $-y$-picture is small, our prediction will be reliable. The situation where $x$ belongs to the "borderland"

$$|(w^* \cdot x) + b^*| \leq 1$$

is more complicated: our algorithm's prediction will depend on the exact positions of the positive and negative examples; to explicate our prediction rule inside this border region is an interesting open problem.

**Remark 5** Transduction is naturally related to a set of algorithms known as instance-based, or case-based learning. Perhaps, the most well-known algorithm in this class is $k$-nearest neighbour algorithm. The transductive algorithm described in this paper, however, is not based on the similarities between examples (as most of the instance-based techniques), but relies on selection of support vectors, and using the support vectors allows us to introduce the confidence and possibility measures.

## 7   COMPUTER EXPERIMENTS

Some experiments for testing the transductive variant of the SV method (described in Section 5 above) have been conducted. We have chosen a simple pattern recognition problem of identifying handwritten digits using a database of US postal data of 9300 digits, where each digit was a $16 \times 16$ vector (cf. LeCun et al. [6]). The experiments were conducted for a subset of these data (800 examples for the training set and 100 examples for the test set), and included a construction of two-class classifier to separate a digit "2" from a digit "7". A set of preliminary experiments showed that the minimum number of errors is achieved with polynomials of degree 2.

The transductive algorithm, described in Section 2, made 5 errors out of 100 examples (twice digit 2 was mistakenly recognised as 7 and three times digit 7 was recognised as 2) with one undecided example. For comparison the results of prediction using the support vector machine show just 1 error (digit 2 was recognised as digit 7).

Our explanation of why support vector machine makes fewer mistakes than our algorithm is as follows. The cases where the new example is not a support vector in one of the pictures are easy and both algorithms made no mistakes. So let us suppose that the new example is a support vector in both pictures. In this case the transductive algorithm, trying to optimize confidence, predicts according to the picture with a larger number of support vectors. But typically it will be the wrong picture that have more support vectors: the power of support vector machines is explained exactly by the fact that real-world data sets can usually be separated with a small number of support vectors. We can see that optimizing confidence and optimizing performance are, at least to some degree, complementary tasks, and we believe that studying the trade-off between them is an interesting direction of future research.

We therefore decided to combine the strength of prediction using the support vector machine with measures of confidence and possibility obtained through our transductive approach (as described in the end of Section 2). The results are presented in Figure 1. Clearly, the data have been split into two clusters: with possibility measure equal to 1 (cluster 1), and with possibility less than 1 (cluster 2). There are 93 correct classifications (denoted by O's) in the first cluster and 5 correct and 1 incorrect (denoted by X's) classifications in the second cluster. Table 1 gives some general characteristics for both clusters.

One of the results that follow from these experiments is that we can assess the quality of the data by using the possibility measure: the new example can be classified with high accuracy when possibility measure is close to 1; and the poor quality data which do not enable us to classify the new example confidently are usually characterised with low measure of possibility.

We are currently applying the described algorithms together with the measures of confidence and possibility to medical diagnostic problem: how to identify the disease (or diseases) for a new patient with certain symptoms given a set of past patients record data. The records were collected at a hospital in Scotland, and our main purpose is to compare the performance of the transductive algorithm with various alternative classifiers (such as those presented in Gammerman and Thatcher [5]).

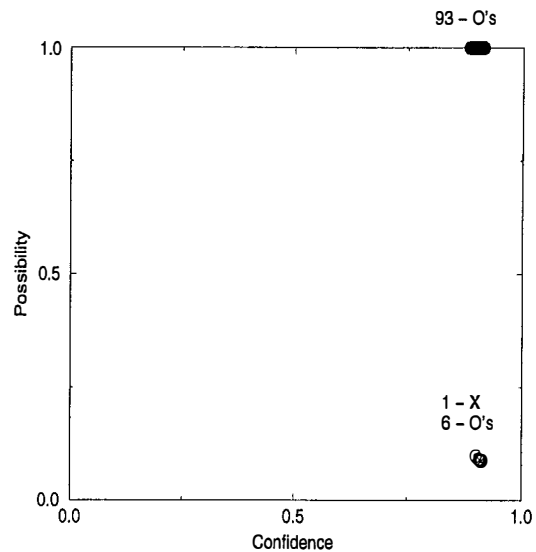

Figure 1: Measures of Confidence and Possibility

The SV method depends on a number of parameters (such as the constant $C$ in (3); it is clear that there are many possible modifications of the SV method: say, we could replace $\xi_i^2$ by $\xi_i^{1+\delta}$, where $\delta > 0$, in (3)). It is important, especially in the transductive variant of the SV method (see Section 5), that the number of the support vectors should be small. We plan to conduct experiments for determining which values of parameters are best in practice. We expect that good

Table 1: Some characteristics of the two clusters (which can be identified by their average possibility) in Figure 1

| CLUSTER | 1 | 2 |
|---|---|---|
| Minimal confidence | 0.883 | 0.901 |
| Maximal confidence | 0.915 | 0.914 |
| Average confidence | 0.902 | 0.910 |
| Average possibility | 1 | 0.0906 |

results will be obtained for the objective function

$$\frac{1}{2}(w \cdot w) + C \sum_i \xi_i^{1+\delta} \to \min,$$

with $C$ large and $\delta > 0$ small (or even $\delta = 0$). The requirement $\delta > 0$ ensures that the objective function is strictly convex; therefore, it is computationally feasible and the arguments of Section 5 apply.

## 8  DISCUSSION

In this section we will very briefly discuss possible directions in which the results of this paper could be developed further.

### 8.1  REGRESSION

An important direction of research is to extend our approach in a computationally feasible way to the problem of regression estimation (see Vapnik [7]). In the latter problem the classifications $y_i$ are no longer required to be binary and can take any real values. In the regression case the key observation (which is an analogue of Lemma 2 above) is the following: if the classifications of the last object in two pictures are more than $2\epsilon$ apart, in at least one of these two pictures the last object will be a support vector. (Here $\epsilon$ is the constant that specifies our tolerance towards inaccurate predictions: deviations of $\epsilon$ or less from the true classification are not punished.) This implies that if the fraction of support vectors is small in all pictures, we will be able to give a prediction with accuracy at most $\epsilon$ and high confidence.

### 8.2  DISTORTION PHENOMENON

The relative number of support vectors is typically small because usually our data are far from being random; for completely random data we can expect that nearly all data points will be support vectors. In our transduction algorithm the incertitude of a correct prediction is determined by the number of support vectors in the "wrong picture", and a natural apprehension is that if that picture is "too wrong", the relative number of support vectors will grow sharply and so the confidence of our prediction will drop. We have not observed this phenomenon in our experiments yet, but there is little doubt that it will be a serious obstacle for very large data sets; in this subsection we discuss a possible remedy.

The value of our permutation measure of impossibility (see (6)) depends on every example $z_i$ only through $z_i$ being a support vector. A natural idea is to use not just whether or not $z_i$ is a support vector, but to take into account the degree of $z_i$'s "supportiveness"; for example, we could use the value of the Lagrange multiplier $\alpha_i$ corresponding to $z_i$. A possible alternative to (6) that will allow us to cope with the distortion phenomenon is

$$p(z_1, \ldots, z_{l+1}) = \frac{f(\alpha_{l+1})(l+1)}{f(\alpha_1) + \cdots + f(\alpha_{l+1})}, \quad (8)$$

where $f$ is some monotonic non-decreasing function with $f(0) = 0$. Sometimes support vectors are defined as the vectors $z_i$ for which $\alpha_i > 0$; under this definition (6) becomes a special case of (8) corresponding to the function $f(\alpha) = \operatorname{sign} \alpha$ (that is, $f(0) = 0$ and $f(\alpha) = 1$ when $\alpha > 0$).

### 8.3  MORE THAN ONE UNCLASSIFIED EXAMPLE

If our task is to predict the classifications $y_{l+1}, \ldots, y_{l+k}$ of $k$ new examples $x_{l+1}, \ldots, x_{l+k}$ given the classifications $y_1, \ldots, y_l$ of the examples $x_1, \ldots, x_l$ in the training set, (8) can be generalized to

$$p(z_1, \ldots, z_{l+1}) = \frac{f(\alpha_{l+1}) + \cdots + f(\alpha_{l+k})}{f(\alpha_1) + \cdots + f(\alpha_{l+k})} \frac{l+k}{k};$$

it is easy to check that this formula defines a valid permutation measure of impossibility.

With each possible prediction

$$y_{l+1} = a_1, \ldots, y_{l+k} = a_k$$

we can associate its incertitude

$$\frac{1}{\min_{(y_{l+1},\ldots,y_{l+k}) \neq (a_1,\ldots,a_k)} p((x_1, y_1), \ldots, (x_{l+k}, y_{l+k}))}$$

and make a prediction with the smallest incertitude.

### 8.4  NON-CONTINUOUS CASE

It is easy (but tedious) to generalize all our results to the case of a probability distribution that is not necessarily continuous; in this subsection we shall only generalize the definition of a permutation measure of impossibility.

A *hyperset in* $Z$ is a subset of $Z$ to each element of which is assigned some *arity* (a positive integer number); the *cardinality* of a hyperset is the sum of the arities of its elements. The *signature* of a finite sequence

$$(z_1, \ldots, z_m) \quad (9)$$

is the hyperset consisting of all elements of (9), with the arity of each element equal to the number of times it occurs in (9).

We let $\mathcal{P}^m(Z)$ stand for the set of all product distributions $P^m$ in $Z^m$ (with $P$ running over all, not necessarily continuous, distributions in $Z$). The following definition of a subclass of $\mathcal{P}^m(Z)$-measures of impossibility is analogous to the definition of permutation measures of impossibility in Section 3. A non-negative measurable function $p : Z^m \to \mathbb{R}$ is an *exchangeable measure of impossibility* if, for any hyperset $b$ of cardinality $m$ in $Z$,

$$\frac{1}{N} \sum_{(z_1,\ldots,z_m) \text{ of signature } b} p(z_1,\ldots,z_m) = 1,$$

where $N$ is the number of all possible sequences $(z_1,\ldots,z_m)$ of signature $b$ (if $b$ assigns arities $b_1,\ldots,b_j$ to its elements, $N = \frac{(b_1+\cdots+b_j)!}{b_1!\ldots b_j!}$).

## 8.5  THE EXCHANGEABILITY MODEL

So far we have assumed that our examples were generated by an i.i.d. source. What we actually used, however, was not this i.i.d. model but a weaker *model of exchangeability*, which only assumes that the examples $z_1,\ldots,z_{l+k}$ are equiprobable. The example of the Bernoulli model shows that the model of exchangeability is strictly weaker than the i.i.d. model (see, e.g, [8]). It is clear that the scheme of Section 4 is "universal" under the exchangeability model, but it remains an open question whether one can make use of the extra strength of the i.i.d. assumption. On the other hand, if the idea of replacing the i.i.d. model by the exchangeability model is to be taken seriously, it would be natural to drop the requirement of measurability in the definitions of a permutation measure of impossibility and exchangeable measure of impossibility (however, this would make little difference in practical applications: in practice we are interested in easily computable, and so *a fortiori* measurable, measures of impossibility).

### Acknowledgments

We thank EPSRC for providing financial support through grant GR/L35812 ("Support Vector and Bayesian Learning Algorithms"). Enlightening comments by the members of Program Committee are gratefully appreciated. We are also grateful to Craig Saunders and Mark Stitson for help with computer experiments.

## References

[1] C. Cortes and V. Vapnik. Support-vector networks. *Machine Learning* 20:1–25, 1995.

[2] A. P. Dawid. Inference, statistical: I. In *Encyclopedia of Statistical Sciences* (S. Kotz and N. L. Johnson, Editors). Wiley, New York, 1983, vol. 4, pp. 89–105.

[3] D. A. S. Fraser. Sequentially determined statistically equivalent blocks. *Ann. Math. Statist.* 22:372, 1951.

[4] A. Gammerman. Machine learning: progress and prospects. Technical Report CSD-TR-96-21, Department of Computer Science, Royal Holloway, University of London, December 1996.

[5] A. Gammerman and A. R. Thatcher. Bayesian diagnostic probabilities without assuming independence of symptoms. *Yearbook of Medical Informatics*, 1992, pp. 323–330.

[6] Y. LeCun, B. Boser, J. S. Denker, D. Henderson, R. E. Howard, W. Hubbard, and L. J. Jackel. Handwritten digit recognition with backpropagation network. *Advances in Neural Information Processing Systems 2*. Morgan Kaufmann, 1990, pp. 396–404.

[7] V. N. Vapnik. *The Nature of Statistical Learning Theory*. Springer, New York, 1995.

[8] V. G. Vovk. On the concept of the Bernoulli property. *Russ. Math. Surv.* 41:247-248, 1986.

[9] V. G. Vovk. A logic of probability, with application to the foundations of statistics (with discussion). *J. R. Statist. Soc.* B 55:317–351, 1993.

[10] V. G. Vovk and V. V. V'yugin. On the empirical validity of the Bayesian method. *J. R. Statist. Soc.* B 55:253–266, 1993.